\documentclass{article}
\usepackage{ijcai23}

\usepackage{times}
\usepackage{soul}
\usepackage{url}
\usepackage[hidelinks]{hyperref}
\usepackage[utf8]{inputenc}
\usepackage[small]{caption}
\usepackage{graphicx}
\usepackage{amsmath}
\usepackage{amsthm}
\usepackage{booktabs}
\usepackage{algorithm}
\usepackage[noend]{algorithmic}
\usepackage[switch]{lineno}

\usepackage{booktabs}
\usepackage{multirow}
\usepackage{stmaryrd}
\usepackage{diagbox}
\usepackage{color, colortbl}
\usepackage{pgf}

\newtheorem{example}{Example}

\newcommand{\name}{\texttt{2L}}
\newcommand{\fullname}{Local Learner}
\newcommand{\microrts}{MicroRTS}
\newcommand{\mc}{\mathcal}   
\newcommand*\GOR{\ |\ }

\title{Choosing Well Your Opponents: \\ How to Guide the Synthesis of Programmatic Strategies}

\author{
Rubens O. Moraes$^{1,2}$
\and
David S. Aleixo$^1$\and
Lucas N. Ferreira$^{2}$\And
Levi H. S. Lelis$^2$
\affiliations
$^1$ Departamento de Inform\'atica, Universidade Federal de Vi\c{c}osa, Brazil\\
$^2$Department of Computing Science, University of Alberta, Canada \\
Alberta Machine Intelligence Institute (Amii)
\emails
rubens.moraes@ufv.br,
david.aleixo@ufv.br,
lnferrei@ualberta.ca, 
levi.lelis@ualberta.ca
}

\begin{document}

\maketitle

\begin{abstract}
    This paper introduces \fullname\ (\name), an algorithm for providing a set of reference strategies to guide the search for programmatic strategies in two-player zero-sum games. Previous learning algorithms, such as Iterated Best Response (IBR), Fictitious Play (FP), and Double-Oracle (DO), can be computationally expensive or miss important information for guiding search algorithms. \name\ actively selects a set of reference strategies to improve the search signal. We empirically demonstrate the advantages of our approach while guiding a local search algorithm for synthesizing strategies in three games, including \microrts, a challenging real-time strategy game. Results show that \name\ learns reference strategies that provide a stronger search signal than IBR, FP, and DO. We also simulate a tournament of \microrts, where a synthesizer using \name\ outperformed the winners of the two latest \microrts\ competitions, which were programmatic strategies written by human programmers. 
\end{abstract}

\section{Introduction}

Programmatic strategies encode game strategies in human-understandable programs. Such programmatic encoding allows domain experts to interpret and modify computer-generated strategies, which can be valuable depending on the application domain (e.g., the games industry). Previous works have used Iterated-Best Response (IBR)~\cite{lanctot2017unified} as the learning algorithm for synthesizing programmatic strategies~\cite{marino2021programmatic}. Given a game, IBR starts with an arbitrary strategy for playing the game and it approximates a best response to it; in the next iteration, it approximates a best response to the best response. This process is repeated a number of iterations and the programmatic strategy synthesized in the last iteration is returned. 

The computation of the best responses in the IBR loop is performed by searching in the programmatic space defined by a domain-specific language. Given a target strategy, the algorithm searches for a program encoding a best response to it. Previous work used local search algorithms for searching in the programmatic space~\cite{marino2021programmatic,medeiros2022can,aleixo23}. The target strategy that IBR provides serves as a guiding function. In the context of local search, when considering the neighbors of a candidate solution, local search algorithms prefer to accept a program that achieves a higher utility value against the target strategy. Since IBR considers a single strategy as a target, the search signal is often weak. This is because the neighbors of a candidate solution that performs poorly against the target strategy are also likely to perform poorly against it---small changes to a losing program will also generate a losing program. Moreover, IBR can loop around the strategy space in games with dynamics similar to Rock, Paper, and Scissors, without making progress toward strong solutions. 

In this paper, we adapt Fictitious Play (FP)~\cite{browniterative} and Double Oracle (DO)~\cite{mcmahan2003planning} to the context of programmatic strategies. FP and DO have been used in the context of neural strategies to overcome some of the weaknesses of IBR~\cite{lanctot2017unified}. Despite providing a better search signal than IBR, we show that FP and DO can still fail to provide relevant information for the search. We then introduce a novel learning algorithm, \fullname\ (\name), that is designed specifically for guiding local search algorithms in the synthesis of programmatic strategies. \name\ uses information gathered while computing best responses to decide the set of target strategies to be used in future iterations of the algorithm as a means of optimizing the search signal. 

We evaluate \name\ on three two-player zero-sum games: \microrts~\cite{ontanon2018first}, Poachers \& Rangers, and Climbing Monkeys. The results show that \name\ synthesized strategies that are never worse and often far superior to strategies synthesized with IBR, FP, and DO in all three domains. We also performed a simulated competition of \microrts\ with strategies synthesized with \name\, IBR, FP, DO, as well as the programmatic strategies that won the last two \microrts\ competitions, which were written by programmers. \name\ obtained the highest average winning rate in our tournament.

\section{Problem Definition} \label{sec:problem_def}

We consider the synthesis of programmatic strategies assuming zero-sum two-player games $\mc{G} = (\mc{P}, \mc{S}, s_{\text{init}}, \mc{A}, \mc{T}, \mc{U})$. Let $\mc{P} = \{i, -i\}$ be the pair of players; $\mc{S}$ be the set of states, with $s_{\text{init}}$ in $\mc{S}$ being the initial state. Each player $i$ can perform an action from a legal set of actions $\mc{A}_i(s)$ in $\mc{A}$ for a given state $s$. 
The action of each player is given by a strategy, which is a function $\sigma_i$ that receives a state $s$ in $\mc{S}$ and returns an action in $\mc{A}_i$ for $s$. 
A transition function $\mc{T}$ receives a state and an action for each player and deterministically returns the next state of the game, which could be a terminal state, where the utility of each player is determined.   
The utility function $\mc{U}$ returns the value of the game in a given state (terminal or not). For $s$, the value of the game is denoted by $\mc{U}(s,\sigma_i,\sigma_{-i})$ when player $i$ follows the strategy $\sigma_i$ and player $-i$, $\sigma_{-i}$. Considering that the game $\mc{G}$ is zero-sum, the utility function for $-i$ is $-\mc{U}(s,\sigma_i,\sigma_{-i})$. In this paper, we encode strategies for $\mc{G}$ as programs written in a domain-specific language (DSL).

\begin{figure}[ht]
\centering
\begin{minipage}{0.16\textwidth}
\includegraphics[width=\linewidth,keepaspectratio=true]{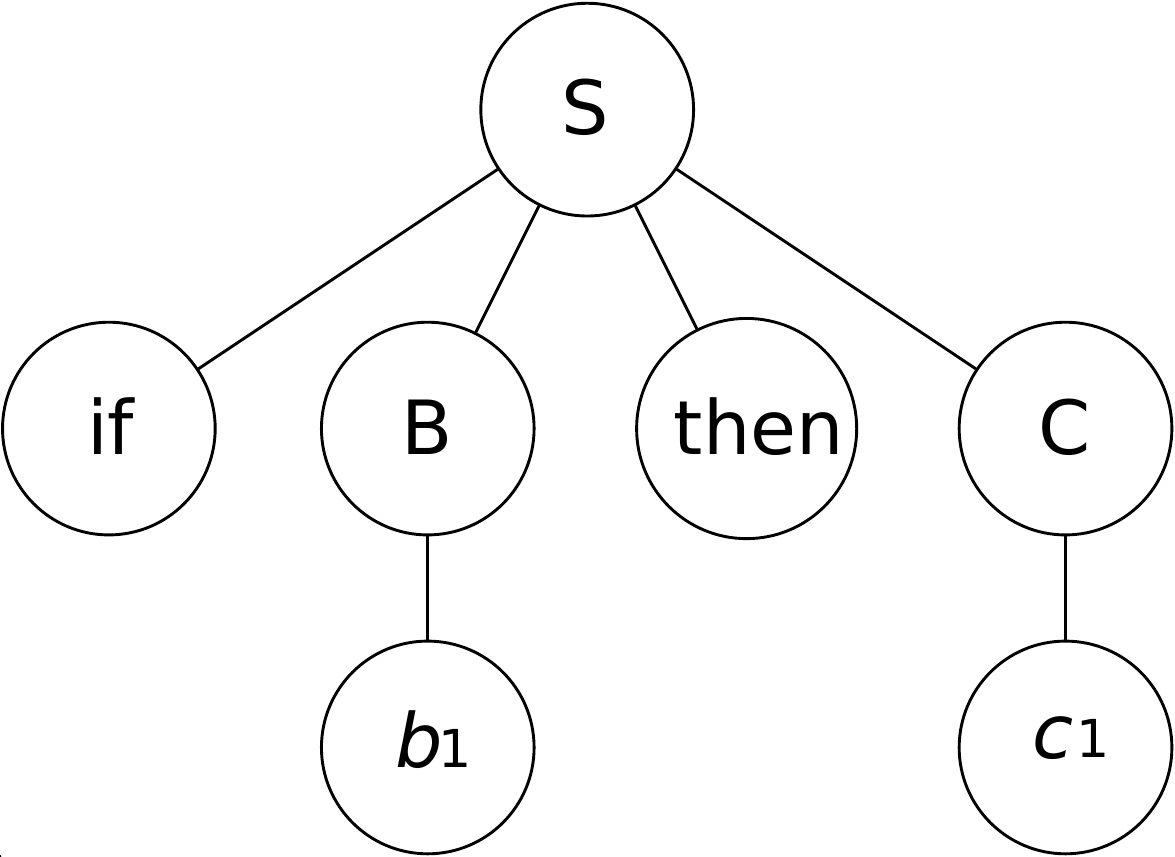}
\end{minipage}
\begin{minipage}{0.23\textwidth}
\begin{align*}
S &\to \, C \GOR \text{if}(B) \text{ then } C \, \\
C &\to c_1 \GOR c_2 \\
B &\to b_1 \GOR b_2 
\end{align*}
\end{minipage}

\caption{AST for ``if $b_1$ then $c_1$'' (left) and DSL in which the program is written (right).}
\label{fig:dsl}
\end{figure}

A DSL can be defined as a context-free grammar $(M, \Omega, R, S)$, where $M$, $\Omega$, $R$, and $S$ are the sets of non-terminals, terminals, relations defining the
production rules of the grammar, and the grammar's initial symbol, respectively. Figure~\ref{fig:dsl} (right) shows an example of a DSL, where $M = \{S, C, B\}$, $\Omega = \{c_1, c_2, b_1, b_2$ if, then$\}$, $R$ are the production rules (e.g., $C \to c_1$), and $S$ is the initial symbol. 

The DSL in Figure \ref{fig:dsl} allows programs with a single command (e.g., $c_1$ or $c_2$) and programs with branching. We represent programs as abstract syntax trees (AST), where the root of the tree is $S$, the internal nodes are non-terminals, and the leaf nodes are terminals. 
Figure~\ref{fig:dsl} (left) shows an example of an AST. We use a DSL $D$ to define the space of programs $\llbracket D \rrbracket$, where each program $p \in \llbracket D \rrbracket$ is a game strategy.

One solves the problem of synthesizing programmatic strategies by solving the following equation
\begin{equation}
\begin{aligned}
\max_{\sigma_{i} \in \llbracket D \rrbracket} \min_{\sigma_{-i} \in \llbracket D \rrbracket} \mc{U}(s_{\text{init}}, \sigma_{i}, \sigma_{-i} ) \,.
\label{eq:syn_optimal}
\end{aligned}
\end{equation}
\noindent

The strategies $\sigma_i$ 
and $\sigma_{-i}$ in $\llbracket D \rrbracket$ able to solve Equation \ref{eq:syn_optimal} define a Nash equilibrium profile in the programmatic space. 
We consider a programmatic variant of PSRO~\cite{lanctot2017unified} to approximate a solution to Equation~\ref{eq:syn_optimal}. 

\section{Programmatic PSRO (PPSRO)}

Let $\lambda$ be a normal-form game defined by $(\Sigma, \mc{P}, U_{\Sigma})$, where $\Sigma = \{\Sigma_i, \Sigma_{-i} \} $ represents a set of strategies for each player in $\mc{P}= \{i, -i\}$, and $U_{\Sigma}$ is the utility payoff table between each pair of strategies in $\Sigma$. A mixed strategy $\sigma$ is a probability distribution over strategies $\Sigma_i$ and $\Sigma_{-i}$ for players $i$ and $-i$, respectively. 
An empirical game of a normal-form game contains only a subset of the strategies of the original game. 

Policy-Space Response Oracles (PSRO) is a framework for learning strategies that ``grow'' an empirical game~\cite{lanctot2017unified}.
In PSRO, the empirical game starts with a single strategy in $\Sigma_i$ and $\Sigma_{-i}$ and it grows these sets by including a new strategy for each player in each iteration of the algorithm. 
Let a mixed strategy over the sets $\Sigma_i$ and $\Sigma_{-i}$ of the empirical game be called a meta-strategy. PSRO grows $\Sigma_i$ and $\Sigma_{-i}$ by adding best responses to meta-strategies. Once a best response is added to a set, a new meta-strategy is computed, and the process is repeated. That is, given a meta-strategy $\sigma_{-i}$ (resp. $\sigma_{i}$), for player $-i$ (resp. $i$), the best response to $\sigma_{-i}$ (resp. $\sigma_{i}$) is added to $\Sigma_i$ (resp. $\Sigma_{-i}$). 

PSRO generalizes algorithms such as IBR, FP, and DO depending on how the meta-strategies are computed. Let $\sigma_k = (p_1, p_2, \cdots, p_n)$ be a meta-strategy for player $k$ ($k$ can be either $i$ or $-i$). Here, $p_j$ in $\sigma_k$ represents the probability in which $\sigma_k$ plays the $j$-th strategy added to the empirical game for player $k$. PSRO generalizes IBR if the meta-strategies are of the form $(0.0, 0.0, \cdots, 1.0)$, i.e., the only strategy in the support of the meta-strategy is the last strategy added to the empirical game. If the meta-strategy $\sigma_{-i}$ with $n$ strategies is of the form $(1/n, 1/n, \cdots, 1/n)$, i.e., all the previous strategies added to the game are played with equal probability, then PSRO generalizes FP.  
PSRO also generalizes DO \cite{mcmahan2003planning} when the meta-strategy is computed by solving the empirical game. 
We use a variant of PSRO, which we call Programmatic PSRO (PPSRO), to approximate a solution to Equation~\ref{eq:syn_optimal}. PPSRO is shown Algorithm~\ref{alg:ppsro}.

\begin{algorithm}[t]
\caption{Programmatic PSRO}
\label{alg:ppsro}
\begin{algorithmic}[1]
\REQUIRE Game $\mc{G}$, DSL $D$, learning algorithm $\Psi$. 
\ENSURE Strategy $\sigma_i$ for player $i$.
\STATE Initialize $\Sigma_i$ and $\Sigma_{-i}$ with one strategy each. \label{ppsro:init}
\STATE Compute utilities $U_\Sigma$ for $(\sigma_i, \sigma_{-i})$
\WHILE{have not exhausted budget}
    \FOR{player $i$ in $\mc{P}$}
        \STATE Compute a meta-strategy $\sigma_{-i}$ with $\Psi(\Sigma,U_\Sigma)$ \label{ppsro:compute_meta}
        \STATE $\sigma'_{i}$ $\gets$ search($\sigma_{i}[-1]$, $\sigma_{-i}$) \label{ppsro:local_search}
        \STATE $\Sigma_i \gets \Sigma_i ~\cup ~\{\sigma'_{i}\}$ \label{ppsro:last_include}
        \STATE Compute entries in $U_\Sigma$ from $\Sigma$
    \ENDFOR
\ENDWHILE
\RETURN Last meta-strategy $\sigma_i$ \label{ppsro:result}

\end{algorithmic}
\end{algorithm}

PPSRO starts by initializing the set of strategies, $\Sigma_i$ and $\Sigma_{-i}$, with two arbitrary strategies (line~\ref{ppsro:init}). 
PPSRO runs a number of iterations according to a given computational budget (e.g., the number of games played). In each iteration, PPSRO invokes a learning algorithm $\Psi$ (e.g., IBR) that receives the current empirical game and returns a meta-strategy $\sigma_{-i}$ (line~\ref{ppsro:compute_meta}). Then it searches in the programmatic space of strategies for a best response $\sigma'_{i}$ to $\sigma_{-i}$. We consider local search algorithms for computing $\sigma'_{i}$. The search algorithm, described in Section~\ref{sec:hill_climbing_synthesis}, initializes its computation with the last strategy added to the empirical game for $i$, which is denoted as $\sigma_i[-1]$ (line~\ref{ppsro:local_search}). The best response $\sigma'_{i}$ is then added to $\Sigma_i$. At the end, PPSRO returns the last meta-strategy $\sigma_i$ as an approximate solution for player $i$ to Equation~\ref{eq:syn_optimal} (line~\ref{ppsro:result}).

The choice of meta-strategies across iterations of PPSRO determines how quickly it is able to approximate a Nash equilibrium profile for the game. Previous work investigated different approaches to define meta-strategies in the context of PSRO and neural policies~\cite{lanctot2017unified,anthony2020learning,muller2020generalized}. However, searching in programmatic space is different from searching in neural space, since the former does not have a gradient signal to guide the search. As we show in our experiments, meta-strategies used with PSRO might not work well with PPSRO.

\section{Hill Climbing for Synthesis of Strategies} \label{sec:hill_climbing_synthesis}

Hill Climbing (HC) is a local search algorithm that starts with an arbitrary candidate solution to a combinatorial search problem and attempts to improve it with greedy changes to the candidate. We use HC to approximate the best responses to strategies $\sigma_{-i}$ in the PPSRO main loop (line~\ref{ppsro:local_search} of Algorithm~\ref{alg:ppsro}). HC receives the last strategy added to the empirical game for player $i$, which is denoted as $\sigma_i[-1]$, and $\sigma_{-i}$. The algorithm returns an approximate best response to $\sigma_{-i}$. This is achieved by searching in the programmatic space defined by the DSL. The starting candidate solution $\sigma_0$ of the search is $\sigma_i[-1]$. HC attempts to approximate a best response to $\sigma_{-i}$ by evaluating neighbor strategies of $\sigma_0$. We update the current candidate solution $\sigma_0$ to a neighbor $\sigma_i'$ if the value $\mc{U}(s_{\text{init}},\sigma_i',\sigma_{-i})$ is greater than $\mc{U}(s_{\text{init}},\sigma_0,\sigma_{-i})$. Otherwise, HC generates and evaluates a new neighbor solution $\sigma_i'$ of $\sigma_0$. This process is repeated until we have exhausted the search budget. HC returns the strategy encountered in search with the highest $\mc{U}$-value as its approximated best response to $\sigma_{-i}$. 

Neighbor solutions are produced by applying a ``mutation'' in the AST of $\sigma_0$. A mutation is carried out by uniformly sampling a non-terminal symbol $S$ in the AST, and replacing the subtree rooted at $S$ with a new subtree. The new subtree is generated by replacing $S$ with the right-hand side of a production rule for $S$ that is selected uniformly at random. The mutation process repeatedly replaces a non-terminal leaf node in the generated program with the right-hand side of a random production rule of the DSL until the program's AST contains only terminal symbols as leaves. 

HC is initialized with a random program only in the first iteration of PPSRO; HC is initialized with the programmatic best response computed in the previous iteration of PPSRO otherwise ($\sigma_i[-1]$ in line~\ref{ppsro:local_search} of Algorithm~\ref{alg:ppsro}).

\section{Shortcomings of Existing Approaches} \label{sec:shortcomings}

The effectiveness of the search algorithm, e.g., HC, for computing a best response depends on the computational cost of $\sigma_{-i}$ and on the information $\sigma_{-i}$ encodes, as we explain next. The meta-strategy $\sigma_{-i}$ determines how fast we can approximate a Nash equilibrium profile for the game. This is because the utility function $\mc{U}(s_{\text{init}}, \sigma_i, \sigma_{-i})$ provides the search signal for the synthesis of a best response $\sigma_{i}$ to $\sigma_{-i}$ in the $\llbracket D \rrbracket$ space. For example, if the meta-strategy $\sigma_{-i}$ with $n$ strategies is of the form $(1/n, 1/n, \cdots, 1/n)$, i.e., all the previous strategies synthesized in the previous iterations are in $\sigma_{-i}$'s support, then $\sigma_{-i}$ is able to provide a richer guiding signal than IBR's meta-strategy, which accounts only for a single strategy. 
Note that PSRO (and PPSRO) with meta-strategies that account for all strategies with equal probability is equivalent to FP~\cite{lanctot2017unified}. Although FP provides a richer search signal, it incurs a higher computational cost as the guiding function $\mc{U}(s_{\text{init}}, \sigma_i, \sigma_{-i})$ requires one to evaluate all strategies in the support of the meta-strategy. Example~\ref{example:ibr} illustrates the IBR's lack of information to guide the search in the game of Poachers and Rangers (P\&R).

P\&R is a simultaneous-move two-player zero-sum game without ties where \textit{rangers} need to protect the gates of a national park to avoid  \textit{poachers} getting inside. In the game, poachers need to attack at least one unprotected gate to enter the park, and rangers succeed if they protect all gates attacked by poachers. Rangers receive the utility of 1 if they protect all attacked gates and -1 otherwise 
The game has a trivial Ranger's dominant strategy, where they protect all the gates. Despite having a trivial solution, the game is particularly hard as a program synthesis task. This difficulty is inherent to the size of the programmatic solution required to solve this game. 
If the number of gates is arbitrarily large, current synthesizers might struggle to synthesize such long programs.
For example, for a game with $n$ gates, the optimal programmatic strategy is any permutation of the instructions in the following program: \texttt{defend}[1], \texttt{defend}[2], $\cdots$, \texttt{defend}[n], which we also denote as \texttt{defend}[1, 2, $\cdots$, n] for conciseness. 

\begin{example}
Let us consider a P\&R instance with 2 gates. In the first iteration, IBR generates an arbitrary strategy for Rangers: \texttt{defend}[2]. In the next iteration, it computes a best response to \texttt{defend}[2]: \texttt{attack}[1]. 
Next, IBR computes a best response to the Poachers strategy, \texttt{attack}[1], so it produces the strategy \texttt{defend}[1]. Then, IBR computes a best response to \texttt{defend}[1], thus generating \texttt{attack}[2] for Poachers. In the next iteration, IBR computes \texttt{defend}[2] as a best response to \texttt{attack}[2]. Note that \texttt{defend}[2] is the strategy in which IBR started the learning procedure---IBR just looped back to the beginning of the process. Since IBR uses only the last synthesized strategy, it can loop over suboptimal strategies which could delay the convergence to the optimal strategy \texttt{defend}[1, 2]. 

By contrast, in FP one considers all previous strategies synthesized in the learning process. Once the empirical game has the strategies \texttt{attack}[1] and \texttt{attack}[2], the search algorithm is guided to synthesize the optimal \texttt{defend}[1, 2].
\label{example:ibr}
\end{example}

DO may strike a balance between computational cost and search guidance, i.e., it includes fewer strategies than FP, but more than IBR in the support of the meta-strategy. With DO, only the strategies in the empirical game that are deemed important, i.e., that are in the support of a Nash equilibrium strategy, will be considered in search. However, DO might still miss important information to guide local search algorithms in the context of PPSRO, as we show in Example~\ref{example:do}.

\begin{example}
Let us consider a P\&R instance with 5 gates. In the first iteration, DO generates two arbitrary strategies:\texttt{defend}[2] and \texttt{attack}[1] for Rangers and Poachers, respectively. Let us assume that PPSRO instantiated as DO generates the empirical game shown in Table~\ref{tbl:matrix_pay_exem} after a few iterations. In the following iteration, PPSRO adds a strategy for Rangers to the empirical game. This is achieved by solving the empirical game shown in Table~\ref{tbl:matrix_pay_exem} to generate a meta-strategy $\sigma_{-i}$ for Poachers and then approximating a best response $\sigma_i$ to $\sigma_{-i}$. The last row of Table~\ref{tbl:matrix_pay_exem} shows the strategy for $-i$ in the Nash equilibrium profile for the empirical game, which is used as the meta-strategy $\sigma_{-i}$. Any strategy $\sigma_i$ for Rangers that defends at least gates $1$, $2$, and $5$ is a best response to $\sigma_{-i}$ since the support of $\sigma_{-i}$ only accounts for \texttt{attack}[1, 2, 5]. The best response $\sigma_i$ does not need to defend the gate $3$, despite being part of the empirical game for Poachers (in strategy \texttt{attack}[1, 2, 3]). If both \texttt{attack}[1, 2, 3] and \texttt{attack}[1, 2, 5] were in the support of $\sigma_{-i}$, PPSRO would be forced to synthesize a strategy that defends gates $1$, $2$, $3$, and $5$. However, DO does not include \texttt{attack}[1, 2, 3] in the support of $\sigma_{-i}$, so PPSRO is only forced to synthesize a strategy that defends gates $1$, $2$, and $5$, which could delay the convergence of the algorithm for missing gate $3$.

\begin{table}[!t]
\setlength{\tabcolsep}{1.3pt}
\begin{tabular}{@{}lccc@{}}
\toprule
\multicolumn{1}{c}{\multirow{2}{*}{\textbf{Rangers}}} & \multicolumn{3}{c}{\textbf{Poachers}}                                                                                   \\
\cmidrule{2-4}
\multicolumn{1}{c}{}                                  & \multicolumn{1}{l}{\texttt{attack}{[}1{]}} & \multicolumn{1}{l}{\texttt{attack}{[}1,2,3{]}} & \multicolumn{1}{l}{\texttt{attack}{[}1,2,5{]}} \\ \midrule
\texttt{defend}{[}2{]}                                       & -1                                  & -1                                      & -1                                      \\
\texttt{defend}{[}1,2,4{]}                                   & \hspace{0.08cm}1                                   & -1                                      & -1                                      \\
\texttt{defend}{[}1,2,3{]}                                   & \hspace{0.08cm}1                                   & \hspace{0.08cm}1                                       & -1                                      \\ \midrule
\multicolumn{1}{c}{$\sigma_{-i}$}    & 0.0                                  & 0.0                                      & 1.0                                     \\ \bottomrule
\end{tabular}
\caption{The empirical game created by DO after a few iterations of the P\&R. This matrix is calculated by evaluating row strategies against column strategies. A score of 1 means that the row strategy beats the column strategy, and -1, the row loses to the column.
}
\label{tbl:matrix_pay_exem}
\end{table}
\label{example:do}
\end{example}

To address these limitations described for IBR, FP, and DO, we propose a new algorithm able to better guide the synthesis of programmatic strategies in the context of PPSRO.

\section{\fullname\ (\name)} \label{sec:NC}

We propose a new instance of PPSRO called \fullname\ (\name), which can overcome the limitations of IBR, FP, and DO presented in the previous section. \name\ defines meta-strategies that are ``in between'' those IBR and FP define in terms of the number of strategies in the meta-strategy's support. \name\ can use more strategies than IBR to provide a better signal to the search algorithm, but it also attempts to use fewer strategies than FP to reduce the computational cost of the evaluation. 
The following P\&R example illustrates how \name\ works. 

\begin{example}
\label{example:local_learner}
Let us consider a P\&R instance with $n > 2$ gates. We initialize 
with an arbitrary strategy  (\texttt{attack}[2]) for Poachers and compute a best response to it: \texttt{defend}[2]. Next iteration, we compute a best response to \texttt{defend}[2]: \texttt{attack}[1]. Next, \name\ returns a meta-strategy $\sigma_{-i}$ for Poachers so we can compute a best response to it and add to the empirical game a new strategy for Rangers. Similarly to what FP would do, in this case, \name\ returns a meta-strategy for Poachers that considers all strategies currently in the empirical game (\texttt{attack}[2] and \texttt{attack}[1]): $\sigma_{-i} = (0.5, 0.5)$. Let us suppose that the search returns the best response \texttt{defend}[1, 2] to $\sigma_{-i}$, which is added to the empirical game. \name\ then returns a meta-strategy $\sigma_i = (0.5, 0.5)$ for Rangers that also considers all strategies currently in the empirical game (\texttt{defend}[2] and \texttt{defend}[1, 2]). While computing a best response to $\sigma_i$, \name\ learns that the strategy \texttt{defend}[2] is redundant and can be dropped from the support of $\sigma_i$ in future iterations. Before finding a best response $\sigma_i$ (e.g., \texttt{attack}[3]), let us assume that the search evaluates strategies \texttt{attack}[1] and \texttt{attack}[2]. Note that \texttt{defend}[2] is a best response to only \texttt{attack}[2], while \texttt{defend}[1, 2] is a best response to both. Given the strategies evaluated in search and that \texttt{defend}[1, 2] is in the support of the meta-strategy, \texttt{defend}[2] does not add new information to the search and can therefore be dropped.  
\end{example}

\name\ initially assumes that all the strategies inserted in the empirical game are helpful in guiding the search, so it adds them to the support of its meta-strategy $\sigma_{-i}$. While computing a best response to $\sigma_{-i}$, it collects data on each strategy in $\sigma_{-i}$ and removes from its support all ``redundant strategies''.

\subsection{Formal Description}
\label{sec:description}

Let $\Sigma_k = \{\sigma_{1,k}, \cdots, \sigma_{n,k}\}$ be the set of strategies for player $k$ in the empirical game 
in an execution of 
PPSRO, where $k$ is either $i$ or $-i$ and $\sigma_{j,k}$ is the $j$-th strategy added for $k$ in the empirical game. Let $\sigma_k = (p_1, \cdots, p_n)$ be a meta-strategy over $\Sigma_k$ where $p_j$ in $\sigma_k$ indicates the probability in which $\sigma_k$ plays the $j$-th strategy in $\Sigma_k$. We denote $p_{j}$ in $\sigma_k$ as $\sigma_k[j]$. 
Let $\Sigma_{\sigma_k}$ be the subset of strategies in the support of $\sigma_k$, i.e., the strategies whose $p_j$-value is greater than zero in $\sigma_k$. 

While computing a best response to a meta-strategy $\sigma_k$, \name\ employs a search algorithm that evaluates a number of strategies as potential best responses to $\sigma_k$. Let $S$ be the set of strategies evaluated in search that are best responded by at least one strategy in $\Sigma_{\sigma_k}$. 
We call \emph{helpful strategies}, denoted $\Sigma_{\sigma_k}^h$, the smallest subset of $\Sigma_{\sigma_k}$ that contains at least one best response to any strategy in $S$. 
We call \emph{redundant strategies} the set $\Sigma_{\sigma_k}$ minus the helpful strategies $\Sigma_{\sigma_k}^h$. 

\begin{example}
In Example~\ref{example:local_learner}, when computing a best response to $\sigma_i = (0.5, 0.5)$ with $\Sigma_{\sigma_i} = \{\texttt{defend}[2], \texttt{defend}[1,2]\}$ we have $S = \{\texttt{attack}[1], \texttt{attack}[2]\}$ and $\Sigma_{\sigma_i}^h = \{\texttt{defend}[1, 2] \}$. \name\ is then able to remove the redundant set $\{\texttt{defend}[2]\}$ from $\Sigma_{\sigma_i}$ for future iterations of the algorithm.
\end{example}

In practice, we are unable to compute the smallest set $\Sigma_{\sigma_k}^h$ possible due to two reasons. First, the search may not find the strategies needed to prove that a strategy is helpful. In Example~\ref{example:local_learner}, if the synthesis algorithm encounters \texttt{attack}[2] but it does not encounter \texttt{attack}[1] during the search, then strategies \texttt{defend}[2] and \texttt{defend}[1, 2] would be ``equally helpful'' and either one could be selected depending on the tie-breaking procedure implemented. Second, finding the smallest set $\Sigma_{\sigma_k}^h$ given $S$ is equivalent to solving a set cover problem, which is NP-hard~\cite{garey1979computers}. \name\ uses a polynomial-time greedy algorithm to approximate a solution to the set cover problem. Namely, we define an initially empty set $S'$. Then, in every iteration, we select the strategy $\sigma$ in $\Sigma_{\sigma_k}$ that is a best response to the largest number of strategies in $S \setminus S'$ and we add all the strategies for which $\sigma$ is a best response to $S'$. We stop when $S = S'$. The strategies selected from $\Sigma_{\sigma_k}$ in this procedure approximate $\Sigma_{\sigma_k}^h$, which gives us an approximation of the redundant strategies.  

\name\ works by executing the following steps:

\begin{enumerate}
\item Initialize $\Sigma_{-i}$ and $\Sigma_{\sigma_{-i}}$  with $\{\sigma_{1,-i}\}$ for some arbitrary strategy $\sigma_{1,-i}$; compute a best response $\sigma_{1, i}$ to $\sigma_{1,-i}$ and initialize $\Sigma_{i}$ and $\Sigma_{\sigma_{i}}$ with $\{\sigma_{1, i}\}$. Define meta-strategies $\sigma_i$ and $\sigma_{-i}$ as $(1.0)$. \label{step1}
\item While there is time for learning and alternating $k$ to be $-i$ in one iteration and $i$ in the next, execute: \label{step2}
\begin{enumerate}
\item Compute a best response $\sigma$ to $\sigma_{-k}$ and add it to $\Sigma_k$ and to $\Sigma_{\sigma_k}$; set $\sigma_{k}[j] = 1.0/|\Sigma_{\sigma_{k}}|$ for all $\sigma_j$ in $\Sigma_{\sigma_{k}}$.\label{best_response}
\item Set $\sigma_{-k}[j] = 0$ and remove it from $\Sigma_{\sigma_{-k}}$ all $\sigma_{j, -k}$ in $\Sigma_{-k}$ that were estimated as redundant. \label{remove_redundant_1}
\item 
Set $\sigma_{-k}[j] = 1.0/|\Sigma_{\sigma_{-k}}|$ for all $\sigma_j$ in $\Sigma_{\sigma_{-k}}$. \label{redistribute_probabilities}
\end{enumerate}

\end{enumerate}
\noindent
\name\ starts by initializing the set of strategies of the empirical game and the set of strategies in the support of the meta-strategy with an arbitrary strategy for one of the players ($-i$ in the pseudocode above). Then, it computes a best response to this arbitrary strategy and uses the best response to initialize $\Sigma_i$ and $\Sigma_{\sigma_i}$. The meta-strategies are of the form $(1.0)$ because the empirical game has a single strategy for each player (see Step~\ref{step1} above). Step~\ref{step2} refers to PPSRO's loop, where it computes the best responses while alternating the players. Once a best response $\sigma$ is computed to strategy $\sigma_{-k}$, it is added to the support of $\sigma_k$ with uniform probability (see Step~\ref{best_response}). 

\name\ estimates which strategies in the support of $\sigma_{-k}$ are redundant while computing the best response $\sigma$ to $\sigma_{-k}$. In Step~\ref{remove_redundant_1}, \name\ removes the redundant strategies from the support of $\sigma_{-k}$ and, in Step~\ref{redistribute_probabilities}, redistributes the probabilities so that each strategy in the support has the same probability. 

\begin{example}
In Example~\ref{example:do}, we showed that DO fails to include both \texttt{attack}[1, 2, 3] and \texttt{attack}[1, 2, 5] in the support of the meta-strategy $\sigma_{-i}$, thus missing the guidance information  \texttt{attack}[1, 2, 3] provides. 
Once the strategy \texttt{attack}[1, 2, 5] is added to the empirical game, the meta-strategy will automatically have both \texttt{attack}[1, 2, 3] and \texttt{attack}[1, 2, 5] in its support. In contrast with DO, \name\ retains both strategies in the support of $\sigma_{-i}$ for the next iteration as long as strategies such as \texttt{defend}[1, 2, 3] and \texttt{defend}[1, 2, 5] are evaluated in search as both \texttt{attack}[1, 2, 3] and \texttt{attack}[1, 2, 5] will be flagged as helpful. 
\end{example}

A weakness of \name\ as presented above is that it can flag as redundant a strategy that is helpful if it does not sample enough strategies in the search. For example, if the meta-strategy for Rangers has both \texttt{defend}[1] and \texttt{defend}[2] in its support, but it never evaluates a strategy that attacks gate 1 in search, then \texttt{defend}[1] will mistakenly be removed from the meta-strategy's support. We implement the following enhancement to fix this weakness. Whenever the search returns a best response $\sigma$ to a meta-strategy $\sigma_{-i}$ (resp. $\sigma_{i}$), we evaluate $\sigma$ against all strategies in the empirical game, including those not in the support of $\sigma_{-i}$ (resp. $\sigma_{i}$). If there is a strategy $\sigma'$ in the empirical game that is a best response to $\sigma$, then it must be that \name\ mistakenly removed $\sigma'$ from the support of the meta-strategy. In this case, we repeat the search for a best response with $\sigma'$ added to the support  of the meta-strategy. 

This enhancement can increase the number of times the search algorithm is invoked in each iteration of the PPSRO loop. While we perform a single search per iteration with IBR, FP, and DO, in the worst case, \name\ can perform a number of searches equal to the number of strategies in the game. This is because, in the worst case, we add all strategies of the empirical game to the support of the meta-strategy. 
Despite the possible additional searches, preliminary experiments showed that this enhancement improves the sampling efficiency of \name. All results in this paper use this enhancement. 

In practice, we do not have the guarantee that the search algorithm used in PPSRO's main loop is able to return a best response to a meta-strategy. So we use whichever approximation the search returns as if it was a best response to the meta-strategy. Moreover, depending on the game, we might not be able to immediately recognize a best response to strategy once we see one, as one would have to prove the strategy to be a best response. This could be problematic, for example, when implementing the enhancement that \name\ re-runs the search if there is a strategy in the empirical game that is a best response to the strategy the search returns. 
We run our experiments in games with utilities of $-1, 0, +1$. 
If a best response cannot be easily verified (e.g., \microrts), then we consider that  $\sigma$ is a best response to $\sigma'$ if $\mc{U}(s_{\text{init}}, \sigma, \sigma') = +1$. 

Once \name\ reaches a computational budget, it can return different strategies as its approximate solution to Equation~\ref{eq:syn_optimal}. Similarly to IBR, it can return the last strategy added to the empirical game for each player. \name\ can also return a mixed strategy that is given by the distribution of strategies added to the empirical game, as does FP. We can also solve the resulting empirical game with linear programming, like DO does, and return the resulting strategy. In this paper, we assume the games have a pure dominant strategy for which IBR's approach of returning the last strategy added to the empirical game is suitable; this is what we use in our experiments. 

\section{Empirical Evaluation} \label{sec:results}

\subsection{Problem Domains} \label{sec:problem_domain}

In addition to P\&R, we introduce Climbing Monkey (CM), another two-player zero-sum game with a trivial optimal strategy that is also challenging in the context of programmatic strategies. In CM, monkeys need to climb to a branch of a tree that is higher than the branch the opponent's monkey is able to reach. The branches need to be climbed one at a time, without skipping any branch. The monkey that climbs to a higher branch wins the game. The game ends in a draw if both monkeys climb to a branch of the same height. For a tree with $n$ branches, a dominant programmatic strategy is $\texttt{climb}[1], \texttt{climb}[2], \cdots, \texttt{climb}[n]$. 
Similarly to P\&R, CM is challenging because, depending on the number of branches, it requires one to synthesize long programs. 

In P\&R, learning algorithms perform better if using a larger number of strategies in the support of meta-strategies as having many strategies helps Rangers converge to a strategy that protects all gates. CM is a game where all one needs to use is the last strategy added to the empirical game, i.e., the strategy that allows the monkey to climb to the highest branch. We hypothesize that \name\ is capable of detecting which strategies are needed in the support of the meta-strategies for these two games.

We also evaluate \name\ in \microrts, a real-time strategy game designed for research. There is an active research community that uses \microrts\ as a benchmark to evaluate intelligent systems.\footnote{https://github.com/Farama-Foundation/MicroRTS/wiki} \microrts\ is a game played with real-time constraints and very large action and state spaces~\cite{lelis2021planning}. Each player can control two types of stationary units (Bases and Barracks) and four types of mobile units (Workers, Ranged, Light, and Heavy). Bases are used to store resources and train Worker units. Barracks can train Ranged, Light, and Heavy units. Workers can build stationary units, harvest resources, and attack opponent units. Ranged, Light, and Heavy units have different amounts of hit points and inflict different amounts of damage to the opponent units. Ranged units differ from each other by causing damage from long distances. In \microrts, a match is played on a grid, which represents the map. Different maps might require different strategies to play the game well.

\subsection{Empirical Methodology}

The games of P\&R and CM allow for a comparison of IBR, FP, DO, and \name\ that is easy to understand and analyze as they have trivial optimal strategies. Experiments with \microrts allow us to compare not only existing learning algorithms with \name, but also other methods for playing \microrts. Namely, we compare the programmatic strategies of IBR, FP, DO, and \name\ with programmatic strategies human programmers wrote to win the last two competitions: COAC\footnote{https://github.com/Coac/coac-ai-microrts} and Mayari.\footnote{https://github.com/barvazkrav/mayariBot} We also include two programmatic strategies that have been used in the \microrts\ competition since 2017: WorkRush (WR) and LightRush (LR). LR was the winner of the 2017 competition. We use seven maps of different sizes: 8$\times$8A BasesWorkers, 
16$\times$16 BasesWorkers, 
24$\times$24A BasesWokers, 24$\times$24 DoubleGame, BWDistantResources 32$\times$32, Chambers 32$\times$32, and 32$\times$32 BasesWorkers. We consider two starting locations (the location of player's base) on each map. When evaluating two strategies, to ensure fairness, each strategy plays an equal number of matches in both locations against the other strategy. 

We are interested in evaluating the sample efficiency of the different approaches, i.e., the strength of the strategies they synthesize as a function of the number of games they need to play to synthesize the strategies. We present plots such as the one in Figure~\ref{fig:CMandPandR}, where the x-axis shows the number of games played and the y-axis a performance metric. We measure performance in P\&R in terms of the number of gates Rangers protect; for CM we measure how high a monkey climbs. 

In the \microrts\ plots (Figure~\ref{fig:hcResultThemselvespt1}) we evaluate the strategy a method returns after a number of games played (x-axis) in terms of its winning rate in a tournament with the strategy the other three methods return at the end of their synthesis process (strategies right side of the plots). In the tournament, each strategy plays the other strategies 10 times, 5 at each starting location on the map. \microrts\ matches can finish in draws. Following previous work, we assign a score of 1.0 for each win and 0.5 for each draw. The winning rate is given by adding the number of wins with half the number of draws, divided by the total number of matches~\cite{ontanon2017combinatorial}. 

Since the mutation operation we use in the hill climbing algorithm is stochastic, we perform multiple independent runs of each experiment and report the average results and standard deviation. The number of runs performed in each experiment is specified below. We use \citeauthor{medeiros2022can} \shortcite{medeiros2022can}'s DSL for \microrts. 
\footnote{Our code is at \url{https://github.com/rubensolv/LocalLearnerIJCAI}}

\subsection{Empirical Results}



\begin{figure}[t]
 \centering
 \includegraphics[width=0.49\textwidth]{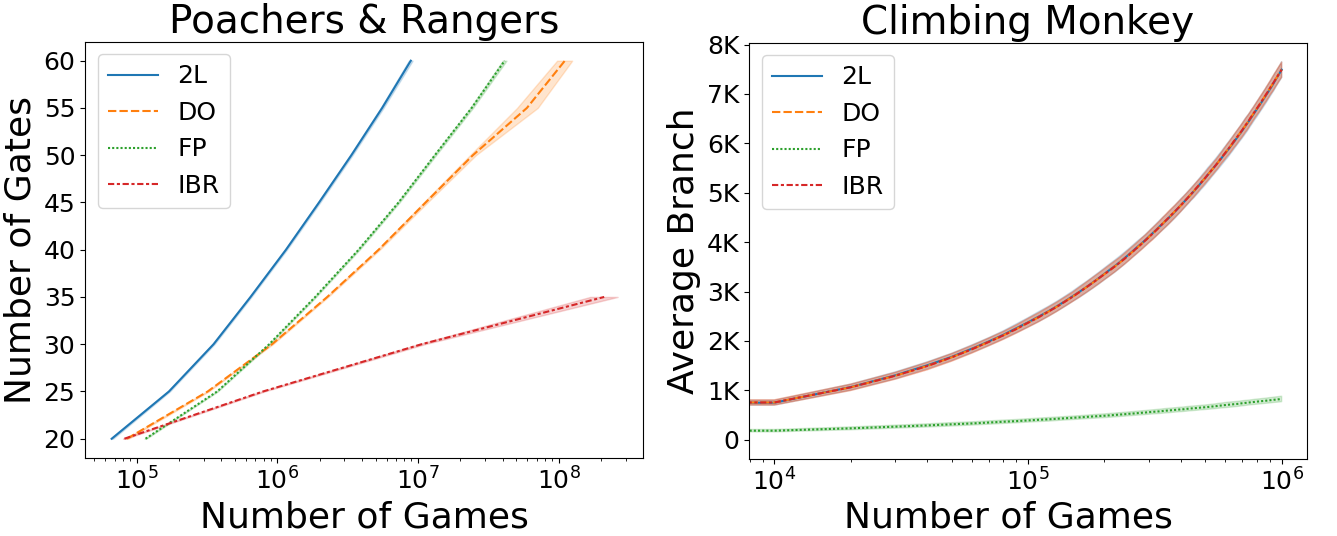}
 \caption{The average number of gates reached on P\&R (left) and the average number of branches on CM (right) for different algorithms.}
 \label{fig:CMandPandR}
 \end{figure}

\begin{figure*}[t]
    \centering
     \includegraphics[width=2.0\columnwidth]{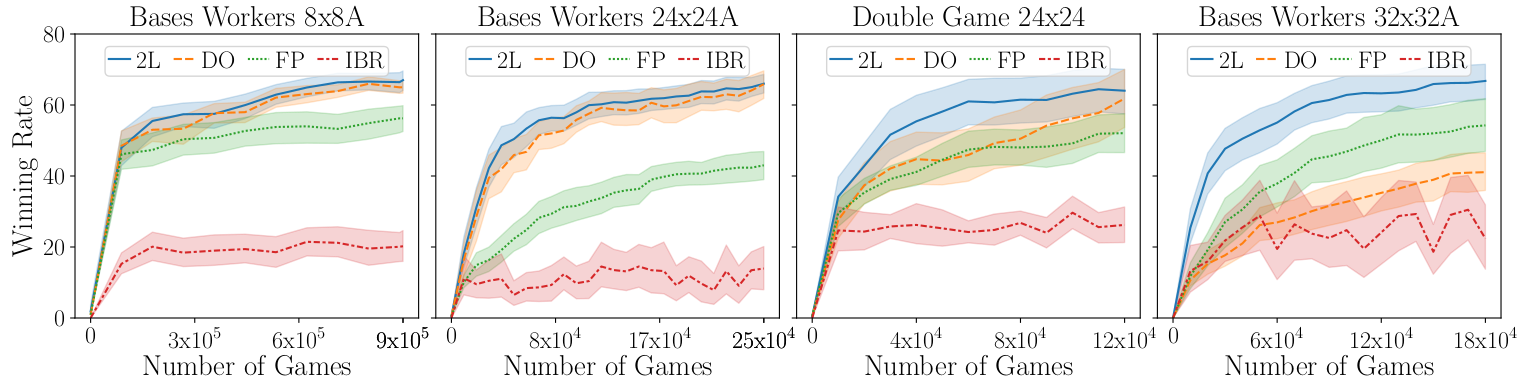}
    \caption{Results of each learning algorithm on four \microrts\ maps using Hill-Climbing local search. These curves represent the learning rate of each algorithm compared to their opponents using the same amount of games.}
    \label{fig:hcResultThemselvespt1}
\end{figure*}

\paragraph{P\&R.} Figure~\ref{fig:CMandPandR} (left) presents the results for P\&R, where each line represents the average number of gates protected over 10,000 independent runs of the algorithms. The x-axis is on a logarithmic scale. \name\ derives strategies that protect more gates with many fewer games played than all other approaches tested. IBR performs the worst likely because it can cycle through strategies it has already seen in learning, as we illustrated in Example~\ref{example:ibr}. FP performs best after \name\ as it is able to remember all previous strategies. However, FP uses more strategies than it needs to make progress, which explains the gap between the \name\ and FP lines. 

\paragraph{CM.} Figure~\ref{fig:CMandPandR} (right) presents the results for CM, where each line is an average of branches climbed over 300 independent runs; the x-axis is in log scale. IBR, DO, and \name\ perform equally well as they all use only the last strategy added to the empirical game in the meta-strategy, which in this domain is the true set of helpful strategies. FP is the worst-performing method in this domain as it unnecessarily uses all strategies from the empirical game in the support of the meta-strategy.

\paragraph{\microrts.} Figure~\ref{fig:hcResultThemselvespt1} presents the results for \microrts, with plots for four representative maps. Each line represents the average of 40 independent runs of each system. \name\ is never worse and is often far superior to the other methods. DO also performs well on most maps, but it is outperformed by a large margin by \name\ and FP in BaseWorkers32x32A. 
IBR performs poorly on all maps. 

\subsection{Simulated Competition Results (\microrts)}

\begin{table*}[t]
\begin{tabular}{@{}lrrrrrrrrr@{}}
\toprule
 \textbf{Method} & \multicolumn{1}{c}{\textbf{LR}} & \multicolumn{1}{c}{\textbf{WR}} & \multicolumn{1}{c}{\textbf{CoaC}} & \multicolumn{1}{c}{\textbf{IBR}} 
 & \multicolumn{1}{c}{\textbf{Mayari}} 
 & \multicolumn{1}{c}{\textbf{FP}} 
 & \multicolumn{1}{c}{\textbf{DO}} & \multicolumn{1}{c}{\textbf{\name}} & \multicolumn{1}{c}{\textbf{Total}} \\
 \midrule
\textbf{LR} & - & 0.71 $\pm$.29 & 0.14 $\pm$.14 & 0.20 $\pm$.15 & 0.00 $\pm$.00 & 0.23 $\pm$.15 & 0.16 $\pm$.18 & 0.13 $\pm$.11 & 0.23 $\pm$.21 \\
\textbf{WR} & 0.29 $\pm$.29 & - & 0.36 $\pm$.36 & 0.25 $\pm$.13 & 0.26 $\pm$.26 & 0.22 $\pm$.10 & 0.14 $\pm$.13 & 0.15 $\pm$.06 & 0.24 $\pm$.15 \\
\textbf{CoaC} & 0.86 $\pm$.14 & 0.64 $\pm$.36 & - & 0.52 $\pm$.21 & 0.09 $\pm$.09 & 0.37 $\pm$.28 & 0.30 $\pm$.26 & 0.24 $\pm$.22 & 0.43 $\pm$.26 \\
\textbf{IBR} & 0.80 $\pm$.15 & 0.75 $\pm$.13 & 0.48 $\pm$.21 & - & 0.49 $\pm$.25 & 0.23 $\pm$.26 & 0.22 $\pm$.26 & 0.22 $\pm$.24 & 0.46 $\pm$.13 \\
\textbf{Mayari} & 1.00 $\pm$.00 & 0.74 $\pm$.26 & 0.91 $\pm$.09 & 0.51 $\pm$.25 & - & 0.36 $\pm$.18 & 0.45 $\pm$.16 & 0.34 $\pm$.19 & 0.62 $\pm$.17 \\
\textbf{FP} & 0.77 $\pm$.15 & 0.78 $\pm$.10 & 0.63 $\pm$.28 & 0.77 $\pm$.26 & 0.64 $\pm$.18 & - & 0.45 $\pm$.29 & 0.41 $\pm$.34 & 0.63 $\pm$.13 \\
\textbf{DO} & 0.84 $\pm$.18 & 0.86 $\pm$.13 & 0.70 $\pm$.26 & 0.78 $\pm$.26 & 0.55 $\pm$.16 & 0.55 $\pm$.29 & - & 0.43 $\pm$.30 & 0.67 $\pm$.15 \\
\midrule
\textbf{\name} & 0.87 $\pm$.11 & 0.85 $\pm$.06 & 0.76 $\pm$.22 & 0.78 $\pm$.24 & 0.66 $\pm$.19 & 0.59 $\pm$.34 & 0.57 $\pm$.30 & - & 0.72 $\pm$.13 \\
\bottomrule
\end{tabular}
\caption{Average winning rate and standard deviation of each learning method (column method) against agents used in the \microrts\ competition. Row values represent the percentage against the column enemy.}
\label{tbl:compe_results_full}
\end{table*}

Table~\ref{tbl:compe_results_full} shows the average results for a set of simulated competitions using the seven maps mentioned in the empirical methodology section.  
Each entry in the table shows the average winning rate and the standard deviation of the row method against the column method; the last column shows the average and standard deviation across a given row. 
The numbers in Table~\ref{tbl:compe_results_full} are the average winning rate computed by simulating 5 tournaments. The strategy we use in each tournament for IBR, FP, DO, and \name is generated as follows. We run each method 8 times, thus producing 8 different strategies each. Then, we run a round-robin evaluation among the 8 strategies of a given method, and the winning strategy in this evaluation is used as the method's strategy in our tournament. For a given tournament, the winning rate is computed by having the strategy of each method play the other strategies 10 times in each map, 5 for each starting location. 


\name\ is the only method to obtain an average winning rate greater than 0.50 against all opponents; \name\ also obtained the highest average winning rate when considering all opponents: 0.72 (column ``Total''). In particular, it obtains average winning rates of 0.76 and 0.66 against COAC and Mayari, respectively, the winners of the two latest competitions. 
A Welch's t-test shows that the difference between \name\ and the competition winners COAC and Mayari, in terms of the total average winning rate, is statistically significant with $p < 10^{-5}$. 

These results on P\&R, CM, and \microrts\ show that \name's approach to defining its meta-strategy can be quite effective in guiding a synthesizer that uses HC search.

\section{More Related Works} \label{sec:related_works}

In addition to PSRO~\cite{lanctot2017unified}, this work is related to programmatic policies~\cite{VermaMSKC18}, where the goal is to synthesize human-readable programs that encode policies to solve reinforcement learning problems~\cite{BastaniPS18,VermaProjection2019}. Generalized planning (GP) is also related because it deals with the synthesis of programs to solve classical planning problems~\cite{FSM-planning,SrivastavaIZZ11,Hu_DeGiacomo_2013,Aguas0J18}. \name\ differs from these works because it learns how to solve two-player games, while the latter focus on single-agent problems.

\citeauthor{marino2021programmatic}~\shortcite{marino2021programmatic,MARINO2022108860} also use local search algorithms to synthesize programmatic strategies, and they also evaluate their system in \microrts. In terms of learning algorithms, they only use IBR, so the IBR version in our experiments is a representation of their work. \citeauthor{medeiros2022can}~\shortcite{medeiros2022can} presents a system for learning sketches with imitation learning as a means of speeding up the computation of programmatic best responses. They focus on the computation of best responses, so their solution can be combined in theory with any of the learning algorithms we evaluated in this paper. 

\section{Conclusions}

In this paper, we introduced Local Learner, a learning algorithm based on the PSRO framework to guide local search algorithms on the task of synthesizing programmatic strategies. \name\ uses information collected from the computation of best responses to approximate a set of helpful strategies to have in the support of \name's meta-strategy, which serves as a guiding function for the search. We empirically showed in three games the advantages of \name\ over adaptations of the learning algorithms IBR, FP, and DO to programmatic strategies. The empirical results show that \name's approach of using information collected during search to determine its own guiding function can be quite effective in practice. \name\ is never worse than the other learning algorithms and is often far superior. In particular, in the game of \microrts, we simulated a competition with the last two winners of the annual \microrts\ competition, and the strategies \name\ synthesized obtained the highest winning rate across all evaluated systems. 

\section*{Acknowledgments}

This research was supported by Canada's NSERC and the CIFAR AI Chairs program and Brazil's CAPES. The research was carried out using computational resources from Compute Canada. We thank the anonymous reviewers for their feedback.




\bibliographystyle{named}
\bibliography{ijcai23}

\end{document}